\documentclass{article}
\usepackage{spconf,amsmath,graphicx}
\usepackage[switch]{lineno}
\usepackage{amssymb}
\usepackage{booktabs}
\usepackage{tabularx}
\usepackage{multirow}   
\usepackage{colortbl}
\usepackage{amssymb}  % 提供 $\checkmark$
\usepackage{pifont}    % 提供 \ding{51}
\usepackage{xcolor}
\usepackage[normalem]{ulem}
\usepackage[colorlinks=true, urlcolor=magenta, linkcolor=blue, citecolor=blue]{hyperref}
\title{TFEC: Multivariate Time-Series Clustering via\\Temporal-Frequency Enhanced Contrastive Learning}
\name{Zexi Tan$^{a}$, Tao Xie$^{a}$, Haoyi Xiao$^{a}$, Baoyao Yang$^{a}$, Yuzhu Ji$^{a}$, An Zeng$^{a}$, Xiang Zhang$^{a}$, Yiqun Zhang$^{a,b,*}$}
  
\address{$^{a}$Guangdong University of Technology, $^{b}$Hong Kong Baptist University}
\begin{document}
%\ninept
%
\maketitle
%

% \linenumbers
\begin{abstract}

Multivariate Time-Series (MTS) clustering is crucial for signal processing and data analysis. Although deep learning approaches, particularly those leveraging Contrastive Learning (CL), are prominent for MTS representation, existing CL-based models face two key limitations: 1) neglecting clustering information during positive/negative sample pair construction, and 2) introducing unreasonable inductive biases, e.g., destroying time dependence and periodicity through augmentation strategies, compromising representation quality. This paper, therefore, proposes a Temporal-Frequency Enhanced Contrastive (TFEC) learning framework. To preserve temporal structure while generating low-distortion representations, a temporal-frequency Co-EnHancement (CoEH) mechanism is introduced. Accordingly, a synergistic dual-path representation and cluster distribution learning framework is designed to jointly optimize cluster structure and representation fidelity. Experiments on six real-world benchmark datasets demonstrate TFEC's superiority, achieving 4.48\% average NMI gains over SOTA methods, with ablation studies validating the design. The code of the paper is available at: \uline{\url{https://github.com/yueliangy/TFEC}}.

\end{abstract}
\begin{keywords}
Time-Series Data, Clustering, Frequency-Domain, Contrastive Learning, Data Augmentation.
\end{keywords}
\section{Introduction}
\label{sec:intro}

Multivariate Time-Series (MTS) clustering~\cite{review15,AAAI23MHCCL}  represents a critical analytical paradigm~\cite{Xie2024ICASSP,ICASSP25ad} with diverse applications in healthcare monitoring~\cite{Bhavani2023Comparison}, industrial diagnostics~\cite{Suh2023Metaheuristic}, and financial analytics~\cite{DUrso2021Trimmed}. Conventional distance-based~\cite{Ma17ICWMW,TCYB20He} and statistical feature methods~\cite{Wang2006Characteristic,WangICDM2007} rely on handcrafted features derived from superficial temporal characteristics (e.g., seasonality/periodicity), yielding shallow representations. These inherently lack the capacity for modeling complex cross-variate dependencies and hierarchical structures while exhibiting pronounced sensitivity to noise regimes and temporal misalignments, fundamentally compromising clustering fidelity in high-dimensional spaces.

% Deep representation learning has consequently emerged to overcome these constraints through autonomous extraction of discriminative embeddings from multivariate temporal structures. Contemporary methodologies diverge into distinct paradigms: reconstruction-based approaches leverage autoencoders to learn latent representations through input reconstruction; generative frameworks employ adversarial or variational architectures to model MTS data distributions; while contrastive learning (CL) explicitly optimizes relative similarity relationships through temporal instance discrimination. This direct instance discrimination enables CL to learn more discriminative and cluster-friendly embeddings than reconstruction or generative approaches, as it bypasses the need to model complex data distributions or irrelevant input details.

Deep representation learning~\cite{Qiu2025Enhancing,ZHANG23TPAMI} overcomes these limitations by autonomously extracting discriminative embeddings from MTS structures. Contemporary methodologies comprise three paradigms: reconstruction-based approaches~\cite{LI2019239,He2022TCYB} employ autoencoders for latent representation learning; generative frameworks~\cite{NEURIPS2019_c9efe5f2,ACM23} utilize adversarial/variational architectures for distribution modeling; while contrastive learning (CL)~\cite{liu2024timesurl,Chang2024ICDE} explicitly optimizes relative similarity through instance discrimination. This direct discrimination enables CL to learn superior discriminative and inherently cluster-friendly embeddings by eliminating the necessity to model complex data distributions or irrelevant input specifics.

% Contrastive learning has emerged as a prominent paradigm for MTS representation learning by optimizing embeddings through positive/negative instance discrimination. It has been shown to possess inherent advantages for cluster structure formation ~\cite{wang2025FCACC}. However, prevailing CL methodologies exhibit two limitations: they predominantly construct contrastive pairs either through latent space manipulations (e.g., temporal dynamics exploitation)~\cite{zhou2023informer,NeurIPS23FOCAL,lee2024soft} or predefined augmentations (e.g., temporal warping/cropping)~\cite{ijcai2021TSTCC,NeurIPS23BU,AAAI23PrimeNet}, thereby neglecting intrinsic cluster-discriminative information within representations during positive/negative sample
% pair construction. Moreover, suboptimal augmentations induce temporal distortions that compromise representation fidelity and continuity.

Contrastive learning has established itself as a powerful paradigm for learning representations from MTS by optimizing embedding similarity through instance discrimination, offering inherent benefits for cluster structure formation~\cite{wang2025FCACC}. Nevertheless, existing CL methods are constrained by two major drawbacks: many approaches either depend extensively on latent-space manipulations, such as those leveraging factorized orthogonal spaces~\cite{NeurIPS23FOCAL}, information-aware augmentations~\cite{zhou2023informer}, or soft assignment strategies~\cite{lee2024soft}; or they construct views based on hand-crafted mechanisms, which include predefined augmentations~\cite{ijcai2021TSTCC, NeurIPS23BU} and pre-training tasks tailored for specific data properties like irregularity~\cite{jiao2020time,AAAI23PrimeNet}. A common shortcoming among these strategies is their frequent oversight of intrinsic cluster-discriminative information during contrastive pair construction. Moreover, such hand-crafted mechanisms, especially suboptimal augmentations, can introduce temporal distortions that compromise the fidelity of the representations.

\begin{figure*}[!t]
\centering
\includegraphics[width=\textwidth]{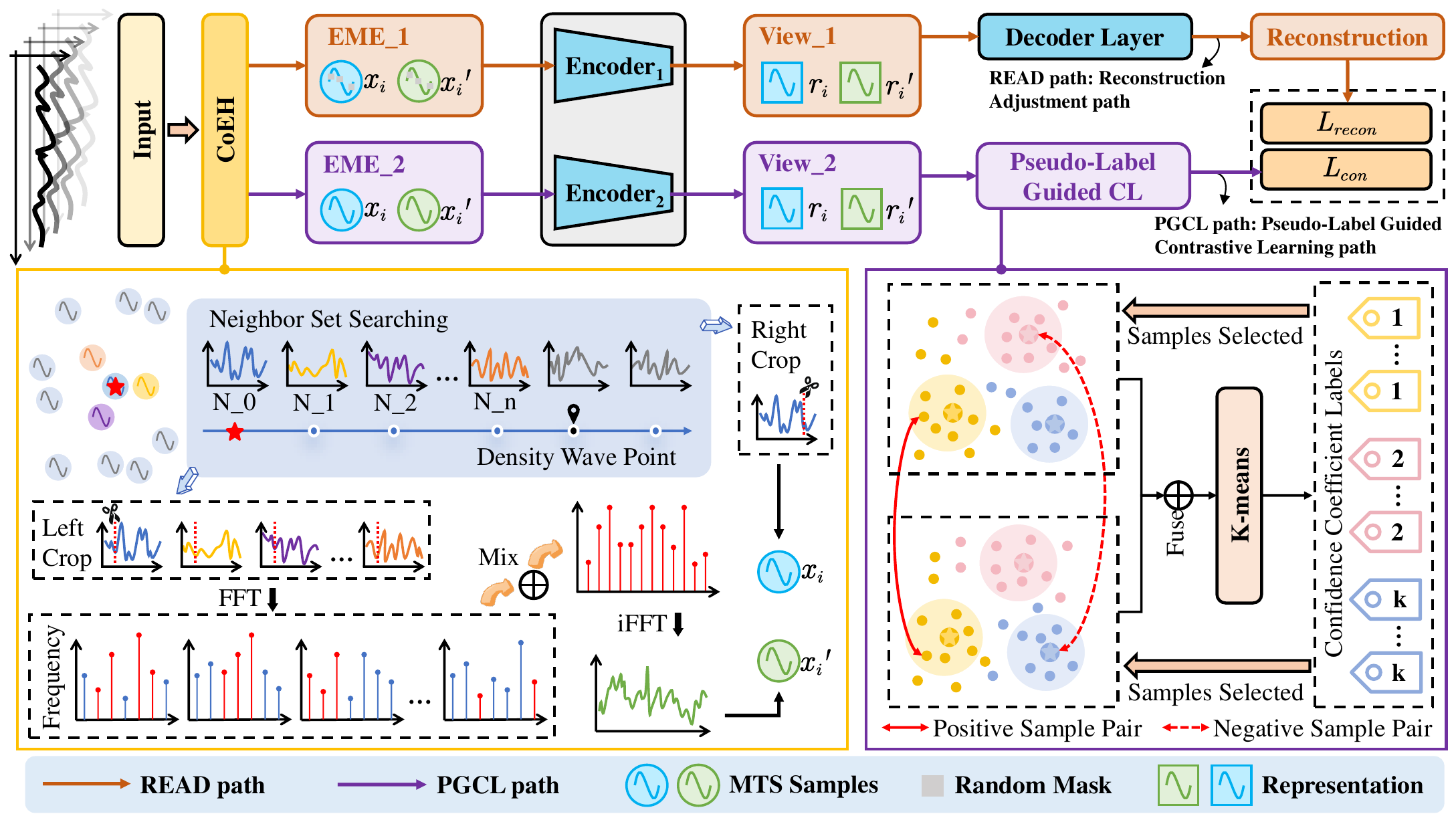}
\caption{\textbf{Overall framework of TFEC}. CoEH generates low-distortion EME. The dual-path architecture processes EME: the PGCL path performs pseudo-label guided contrastive learning on cluster structures based on high-confidence samples, while the READ path stabilizes representations via mask reconstruction.
}
\label{framework}
\end{figure*}

To address these challenges, this paper proposes TFEC: a Temporal-Frequency Enhanced Contrastive learning framework. Our core innovation integrates: (1) A temporal-frequency Co-EnHancement (CoEH) mechanism generating low-distortion representations through adaptive spectral mixing while preserving temporal topology, and (2) A synergistic dual-path architecture where cluster-aware contrastive learning explicitly leverages inherent categorical structures for discriminative embedding formation, while reconstruction-based stabilization maintains representation integrity. This unified paradigm establishes an end-to-end self-supervised framework for optimized MTS clustering, effectively mitigating temporal distortions while fully exploiting underlying cluster structures to improve representation quality. The three main contributions of TFEC are summarized as follows:

% \begin{itemize}
% \item \textbf{A Temporal-Frequency Co-Enhancement Mechanism}: This paper proposes a dual-domain enhancement scheme that preserves temporal continuity through aligned cropping while performing adaptive frequency mixing with proximate neighbors. This can generate low-distortion Enhanced MTS Dataset (EMD).
% \item \textbf{Cluster-Aware Contrastive Learning with High-Confidence Sampling}: This work develops a pseudo-label guided paradigm that fuses dual-view EMD embeddings to generate cluster assignments, then selects high-confidence intra-cluster exemplars for contrastive pairing. This explicitly leverages representation-inherent clustering structures to construct reliable positive (same-cluster) and negative (cross-cluster) pairs.
% \item \textbf{A Synergistic Dual-Path Representation and Cluster Distribution Optimization Framework}: The Pseudo-Label Guided CL (PGCL) path and the REconstruction ADjustment (READ) path together form a complementary dual-path, responsible for optimizing cluster distribution and ensuring the efficacy of representation, respectively, to improve the clustering performance of the model.
% \end{itemize}

\begin{itemize}
\item \textbf{Temporal-Frequency Co-Enhancement Mechanism}: This paper introduces a dual-domain co-enhancement strategy that preserves temporal coherence through aligned cropping and enriches representations via adaptive frequency mixing with semantically proximate neighbors, enabling the generation of a low-distortion Enhanced MTS Embedding (EME) with improved discriminability and fidelity.

\item \textbf{Cluster-Aware Contrastive Learning with High-Confidence Sampling}: A pseudo-label guided paradigm is proposed to fuse dual-view representations for cluster assignment, followed by the selection of high-confidence intra-cluster samples to form reliable contrastive pairs. This approach explicitly harnesses inherent cluster structures, substantially improving the quality and stability of learned representations.

\item \textbf{Synergistic Dual-Path Learning Architecture}: This paper designs a dual-path architecture comprising a Pseudo-label Guided Contrastive Learning (PGCL) path and a REconstruction ADjustment (READ) path. The PGCL path refines cluster compactness and separation, while the READ path enhances representation fidelity through masked EME reconstruction. Their synergistic interaction ensures discriminative and cluster-friendly representations.
\end{itemize}

% framework.

\section{Methodology}
As shown in Fig.~\ref{framework}, the TFEC framework addresses the challenges of temporal distortion and underutilized clustering information through two core innovations: (1) A \textbf{Temporal-Frequency Co-Enhancement Mechanism} for generating low-distortion EME, and (2) A \textbf{Synergistic Dual-Path Learning Architecture} for learning discriminative and low-distortion representations. 

%%%%%%%%%%%%%%%%%%%%%%%%%%%%%%%%%数据增强部分

\subsection{Temporal-Frequency Co-Enhancement Mechanism}
Given a MTS dataset $\mathcal{X} = \{\mathbf{x}_1, \mathbf{x}_2, \dots, \mathbf{x}_N\}$ where each $\mathbf{x}_i \in \mathbb{R}^{T \times F}$, this paper first preserves temporal integrity through aligned cropping. This ensures that local temporal structures and phase coherence are maintained, avoiding the introduction of spurious inductive biases.

To further enhance the embeddings without introducing unreasonable inductive biases, this paper incorporates frequency-domain information from semantically proximate neighbors. These neighbors are identified via a density-aware selection process that ensures physically meaningful associations, avoiding arbitrary or noisy pairings. For each sample $\mathbf{x}_i$, this work applies the Fast Fourier Transform (FFT) to its temporally aligned segments:
\begin{equation}
\mathbf{q}^L_i = \mathcal{F}(x^L_i), \quad \mathbf{q}^L_{\langle i,p \rangle} = \mathcal{F}(x^L_{\langle i,p \rangle}),
\end{equation}
where $x^L_i$ and $x^L_{\langle i,p \rangle}$ denote cropped segments of the processed MTS and its neighbor, respectively. The frequency embeddings are then adaptively mixed using importance weights $\delta_p$ derived from feature-space distances, resulting in a blended frequency embedding:
\begin{equation}
\mathbf{F} = \mathbf{q}^L_i + \sum_p \delta_p \cdot \mathbf{q}^L_{\langle i,p \rangle}.
\end{equation}
Finally, the inverse FFT is applied to synthesize the enhanced time-series sample, preserving both temporal structure and enriched frequency characteristics. This dual-domain strategy effectively augments the data while minimizing distortion, facilitating the learning of more robust representations.

%%%%%%%%%%%%%%%%%%%%%%%%%%%%%%%%%%%%%%%Dual-path协同训练部分
\subsection{Synergistic Dual-Path Learning Architecture}
The dual-path architecture jointly improves representation stability and cluster distribution fidelity. The READ path processes EME which has been masked randomly through autoencoding to mitigate training drift. Concurrently, the PGCL path extracts discriminative encodings $\mathbf{r}, \mathbf{r}'$ from dual representation views. Cluster assignments emerge from fused representations $\mathbf{R} = (\mathbf{r} + \mathbf{r}')/2$ via $K$-means initialization. Crucially, high-confidence samples are selected using cluster-aware pseudo-labels:
\begin{equation}
\text{CONF}_i = \exp\left({-\|\mathbf{R}_i - \mathbf{c}_p\|^2}\right),
\end{equation}
where $\mathbf{c}_p$ denotes the $p$-th cluster centroid. This confidence metric quantifies representation-cluster alignment reliability, enabling the construction of unambiguous contrastive pairs. Positive pairs ($\mathcal{P}$) comprise same-cluster high-confidence samples from different views, while negative pairs ($\mathcal{N}$) are formed between distinct central samples of high-confidence samples. This strategy explicitly leverages inherent cluster structures rather than relying on arbitrary latent space manipulations, ensuring contrastive signals align with genuine categorical boundaries. The READ path's reconstruction stability and PGCL's cluster distribution improvement form complementary objectives that mutually reinforce representation quality.

\begin{table}[!t]
\caption{Statistics of datasets. }\label{tbl_summary}
\centering
\resizebox{\linewidth}{!}{\begin{tabular}{c|c|ccc|c}
 \toprule
            No. & Datasets & \#$N$ & \#$T$ & \#$F$ & \# Classes \\ 
            \midrule
            1 &  AtrialFibrillation (UEA 2018)  & 15 & 640 & 2 & 3 \\ 
            2 &  ERing (UEA 2018)  & 30 & 65 & 4 & 6 \\ 
            3 &  RacketSports (UEA 2018)  & 152 & 30 & 6 & 4 \\ 
            4 &  Libras (UEA 2018)  & 180 & 45 & 2 & 15 \\ 
            5 &  StandWalkJump (UEA 2018)  & 15 & 2500 & 4 & 3 \\ 
            6 &  NATOPS (UEA 2018)  & 180 & 51 & 24 & 6 \\ 
            \bottomrule
\end{tabular}}
\end{table}

\subsection{Loss Function and Model Training}
The hybrid loss function synergistically integrates contrastive learning and reconstruction objectives. Pseudo-label guided contrastive loss minimizes intra-cluster variance while maximizing inter-cluster separation:
\begin{equation}
\mathcal{L}_{\text{con}} = \underbrace{\frac{1}{|\mathcal{P}|} \sum_{(i,j) \in \mathcal{P}} \|\mathbf{r}_i - \mathbf{r}'_j\|^2}_{\text{Positive alignment}} + \alpha \underbrace{\frac{1}{|\mathcal{N}|} \sum_{(p,q) \in \mathcal{N}} \frac{\langle \mathbf{c}_p, \mathbf{c}_q \rangle}{\|\mathbf{c}_p\|\|\mathbf{c}_q\|}}_{\text{Negative separation}},
\end{equation}
where $\alpha$ balances attraction and repulsion forces. Simultaneously, the READ path computes the reconstruction loss $\mathcal{L}_{\text{recon}}$ for the reconstructed MTS. The joint optimization objective integrates both components through a trade-off hyper-parameter $\beta$:
\begin{equation}
\mathcal{L}_{\text{total}} = \beta \mathcal{L}_{\text{con}} + (1-\beta)\mathcal{L}_{\text{recon}}.
\end{equation}
This formulation enables end-to-end training where reconstruction stabilizes feature learning while contrastive signals sharpen cluster boundaries. The dual-path learning ensures representations retain temporal fidelity while achieving separability, fulfilling critical requirements for effective MTS clustering. The Adam optimizer minimizes $\mathcal{L}_{\text{total}}$ through gradient updates that progressively refine both cluster distribution and representation efficacy.

% \begin{figure}[!t]
% \centering
% \includegraphics[width=\columnwidth]{accuracy_com.pdf}
% \caption{\textbf{Clustering performance comparison on six UCR datasets.}}
% \label{performance}
% \end{figure}

\section{Experiment}
% \textbf{Experimental Settings.} To rigorously evaluate TFEC's performance, we conduct comparative and ablation studies using \textbf{six diverse UCR Archive datasets}~\cite{bagnall2018UCRmultivariatetimeseries}: \textit{Beef}, \textit{Coffee}, \textit{Adiac}, \textit{ArrowHead}, \textit{BME} and \textit{Car}. As shown in Table~\ref{tbl_summary}, these datasets represent varied MTS characteristics including variable lengths ($T \in [128, 577]$) and types of labels ($Labels \in [2, 37]$). We benchmark against \textbf{four SOTA deep clustering methods}: \textit{Times-URL}~\cite{liu2024timesurl} , \textit{UNITS}~\cite{NEURIPS2024UniTS} , \textit{DropPatch}~\cite{Qiu2025Enhancing}, and \textit{FCACC}~\cite{wang2025FCACC} , with \textit{k-means}~\cite{MacQueen1967SomeMF} as the traditional baseline. Performance is quantified through \textbf{three established metrics}: Clustering Accuracy (ACC) measuring label alignment, Normalized Mutual Information (NMI) assessing cluster purity, and F1-Score (F1) evaluating clustering consistency between predicted and ground-truth clusters. All the experiments are coded using Python 3.10. Due to space constraints, comprehensive results and more details are available at \url{https://github.com/yueliangy/TFEC}.

\textbf{Experimental Settings.} This paper evaluates TFEC comprehensively through \textbf{five types of experiments}: \textit{Clustering Performance Evaluation}, \textit{Ablation Study}, \textit{Significance Test}, \textit{Comparison with Other Augmentation Mechanisms} and \textit{Efficiency Evaluation}. Experiments are conducted on \textbf{six diverse UEA datasets}~\cite{bagnall2018ueamultivariatetimeseries}: \textit{AtrialFibrillation}, \textit{ERing}, \textit{RacketSport}, \textit{Libras}, \textit{StandWalkJump} and \textit{NATOPS}. As shown in Table~\ref{tbl_summary}, these datasets exhibit varied lengths ($T \in [30, 2500]$) and label types ($Labels \in [3, 15]$). This paper compares against \textbf{four SOTA deep clustering methods}: \textit{TimesURL}~\cite{liu2024timesurl}, \textit{UNITS}~\cite{NEURIPS2024UniTS}, \textit{DropPatch}~\cite{Qiu2025Enhancing}, and \textit{FCACC}~\cite{wang2025FCACC}, with \textit{K-means}~\cite{MacQueen1967SomeMF} as baseline. Performance is measured using \textbf{three metrics}: Clustering Accuracy (ACC), Normalized Mutual Information (NMI), and F1-Score (F1). All experiments are implemented in Python 3.10. Due to space constraints, full results are available as online \uline{\href{https://github.com/yueliangy/TFEC-Appendix/blob/main/ICASSP'26_TFEC_Appendix.pdf}{Appendix}}.

\begin{figure}[!t]
\centering
\includegraphics[width=\columnwidth]{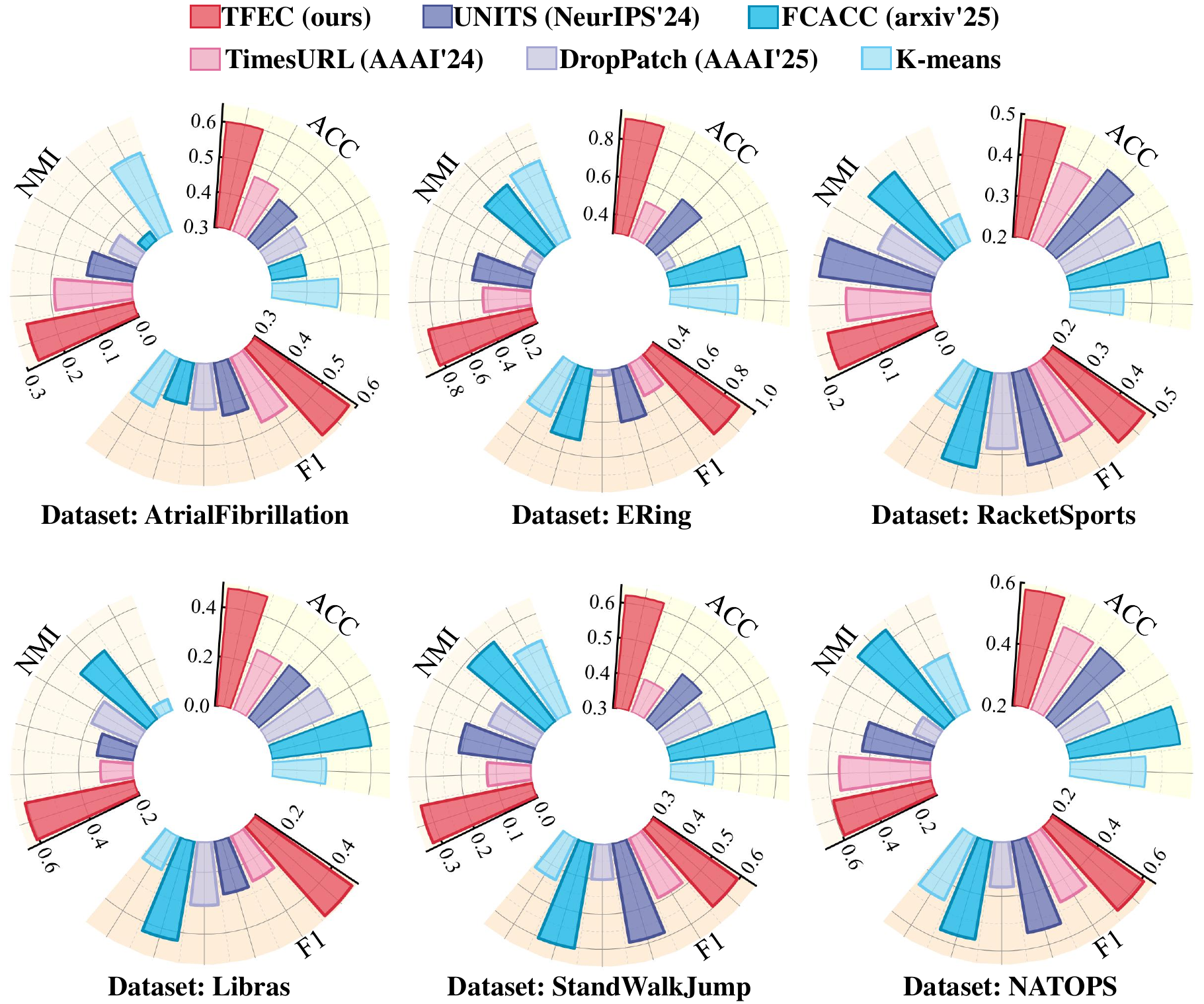}
\caption{Clustering performance comparison on six UEA datasets using ACC, F1 and NMI as evaluation metrics.}
\label{performance}
\end{figure}

% \begin{figure}[ht!]
% \centering
% \includegraphics[width=0.5\textwidth]{Ablation_ACC_new.pdf}
% \caption{Comparison of ablated versions of TFDC.}
% \label{Ablation}
% \end{figure}

% \textbf{Clustering Performance Evaluation (Fig.~\ref{performance}).} 
% TFEC demonstrates superiority across all six datasets; particularly noteworthy are the substantial performance gains on Adiac and Coffee against the strongest baseline, where the CoEH effectively preserves structural integrity while enriching discriminative features. The framework maintains robust performance on long-duration sequences (Beef) and complex multi-class problems (Car), confirming its adaptability to diverse temporal characteristics. The comparatively marginal advantage observed on ArrowHead may be attributed to its simpler categorical structure, where even conventional methods can achieve satisfactory performance. 
% %These comprehensive results validate TFEC's capability in generating superior, low-distortion representations that establish new SOTA performance, significantly outperforming existing deep learning benchmarks.
% In summary, TFEC is capable of generating low-distortion representations, significantly outperforming SOTAs in clustering performance.

\textbf{Clustering Performance Evaluation (Fig.~\ref{performance}).}
TFEC consistently achieves superior performance across all six datasets, with particularly notable gains on \textit{AtrialFibrillation} and \textit{ERing}, where it substantially outperforms the strongest baselines. The CoEH mechanism effectively preserves structural integrity while enriching discriminative features, contributing to these improvements. The framework also maintains robust performance on long-duration sequences such as \textit{StandWalkJump} and complex multi-class datasets like \textit{Libras}, demonstrating its adaptability to diverse temporal characteristics. Although the advantage is relatively narrower on datasets like \textit{RacketSports} and \textit{NATOPS}, TFEC still achieves the highest scores across all metrics, confirming its ability to produce low-distortion, cluster-friendly representations that significantly surpass existing SOTA methods.

% \textbf{Ablation Study (Table~\ref{tb:ablation}).}
% The study validates the necessity of all three modules (CoEH, PGCL, READ). Removing any module generally degrades performance. Exceptions occur on ArrowHead, where PGCL removal slightly improves NMI due to limited contrastive benefits in discriminative spaces, and on Car, where retaining both CoEH and READ compensates for PGCL’s absence. Notably, ablating both PGCL and READ paths causes the most significant performance drop, underscoring their complementary roles. The full TFEC achieves optimal performance via synergistic temporal-frequency enhancement and dual-path learning.

\textbf{Ablation Study (Table~\ref{tb:ablation}).}
The study confirms the necessity of all three core modules (CoEH, PGCL, READ), as removing any component generally degrades performance. Exceptions include \textit{RacketSports}, where ablating PGCL yields a marginal NMI gain, suggesting limited contrastive benefits in its discriminative space; and \textit{NATOPS}, where excluding either PGCL or READ maintains stability. Notably, jointly ablating PGCL and READ causes the most severe performance drop, underscoring their complementary roles. The full TFEC achieves optimal performance via synergistic temporal-frequency enhancement and dual-path learning.

\textbf{Other Experiments (See the \uline{\href{https://github.com/yueliangy/TFEC-Appendix/blob/main/ICASSP'26_TFEC_Appendix.pdf}{Appendix}}).} The significance test confirms the statistical superiority of TFEC. The temporal-frequency co-enhancement consistently outperforms five common augmentation strategies in preserving temporal integrity and enriching discriminative features. Efficiency Evaluation shows that TFEC maintains competitive and scalable performance as data dimensions grow.

\begin{table}[!t]
\centering
% \caption{Ablation Study. ``$\checkmark$'' indicates that the corresponding key component is included. The symbol ``\textcolor{red}{\ddag}'' represents the expected performance degradation of the ablated versions compared to the full TFEC.}

\caption{Ablation study. Checkmarks ``$\checkmark$'' denote included components. The red dagger ``\textcolor{red}{\ddag}'' indicates performance degradation relative to the complete TFEC model.}

\label{tb:ablation}
\small
\setlength{\tabcolsep}{2.5pt}
\begin{tabular}{c|ccc|c|c|c}
\toprule
\multirow{2}{*}{Datasets} & \multicolumn{3}{c|}{Key Components} & \multicolumn{3}{c}{Evaluation Metrics} \\
\cmidrule(lr){2-4} \cmidrule(lr){5-7}
                         & CoEH & PGCL & READ & ACC & F1 & NMI \\  
\midrule
\multirow{5}{*}{AtrialFibrillation}   & $\checkmark$ & $\checkmark$ & $\checkmark$ & 0.6000\textcolor{white}{\ddag} & 0.5841\textcolor{white}{\ddag} & 0.2661\textcolor{white}{\ddag} \\
                        & & $\checkmark$ & $\checkmark$ & 0.5556\textcolor{red}{\ddag} & 0.5322\textcolor{red}{\ddag} & 0.2646\textcolor{red}{\ddag} \\
                          & $\checkmark$ & & $\checkmark$ & 0.4889\textcolor{red}{\ddag} & 0.4462\textcolor{red}{\ddag} & 0.1926\textcolor{red}{\ddag} \\
                          & $\checkmark$ & $\checkmark$ & & 0.5556\textcolor{red}{\ddag} & 0.5136\textcolor{red}{\ddag} & 0.2800\textcolor{white}{\ddag} \\ 
                          & $\checkmark$ & & & 0.4889\textcolor{red}{\ddag} & 0.4336\textcolor{red}{\ddag} & 0.2086\textcolor{red}{\ddag} \\ \midrule

\multirow{5}{*}{ERing}   & $\checkmark$ & $\checkmark$ & $\checkmark$ & 0.9062\textcolor{white}{\ddag} & 0.9068\textcolor{white}{\ddag} & 0.8339\textcolor{white}{\ddag} \\
                        & & $\checkmark$ & $\checkmark$ & 0.8938\textcolor{red}{\ddag} & 0.8950\textcolor{red}{\ddag} & 0.8209\textcolor{red}{\ddag} \\
                          & $\checkmark$ & & $\checkmark$ & 0.8519\textcolor{red}{\ddag} & 0.8852\textcolor{red}{\ddag} & 0.8272\textcolor{red}{\ddag} \\
                          & $\checkmark$ & $\checkmark$ & & 0.8926\textcolor{red}{\ddag} & 0.8932\textcolor{red}{\ddag} & 0.8226\textcolor{red}{\ddag} \\
                          & $\checkmark$ & & & 0.6593\textcolor{red}{\ddag} & 0.6320\textcolor{red}{\ddag} & 0.6795\textcolor{red}{\ddag} \\ \midrule

\multirow{5}{*}{RacketSports}   & $\checkmark$ & $\checkmark$ & $\checkmark$ & 0.4868\textcolor{white}{\ddag} & 0.4758\textcolor{white}{\ddag} & 0.1737\textcolor{white}{\ddag} \\
                        & & $\checkmark$ & $\checkmark$ & 0.4583\textcolor{red}{\ddag} & 0.4416\textcolor{red}{\ddag} & 0.2082\textcolor{white}{\ddag} \\
                          & $\checkmark$ & & $\checkmark$ & 0.4561\textcolor{red}{\ddag} & 0.4430\textcolor{red}{\ddag} & 0.1808\textcolor{white}{\ddag} \\
                          & $\checkmark$ & $\checkmark$ & & 0.4189\textcolor{red}{\ddag} & 0.4330\textcolor{red}{\ddag} & 0.2105\textcolor{white}{\ddag} \\
                          & $\checkmark$ & & & 0.3311\textcolor{red}{\ddag} & 0.3115\textcolor{red}{\ddag} & 0.0483\textcolor{red}{\ddag} \\ \midrule
                            
\multirow{5}{*}{Libras}   & $\checkmark$ & $\checkmark$ & $\checkmark$ & 0.4778\textcolor{white}{\ddag} & 0.4886\textcolor{white}{\ddag} & 0.6074\textcolor{white}{\ddag} \\
                        & & $\checkmark$ & $\checkmark$ & 0.4537\textcolor{red}{\ddag}& 0.4759\textcolor{red}{\ddag} & 0.5895\textcolor{red}{\ddag} \\
                          & $\checkmark$ & & $\checkmark$ & 0.4796\textcolor{white}{\ddag} & 0.4838\textcolor{red}{\ddag} & 0.5884\textcolor{red}{\ddag} \\
                          & $\checkmark$ & $\checkmark$ & & 0.4537\textcolor{red}{\ddag} & 0.4759\textcolor{red}{\ddag} & 0.5895\textcolor{red}{\ddag} \\
                          & $\checkmark$ & & & 0.2185\textcolor{red}{\ddag} & 0.2220\textcolor{red}{\ddag} & 0.2465\textcolor{red}{\ddag} \\ \midrule
                            
\multirow{5}{*}{StandWalkJump}   & $\checkmark$ & $\checkmark$ & $\checkmark$ & 0.6222\textcolor{white}{\ddag} & 0.5961\textcolor{white}{\ddag} & 0.3231\textcolor{white}{\ddag} \\
                        & & $\checkmark$ & $\checkmark$ & 0.5778\textcolor{red}{\ddag} & 0.5566\textcolor{red}{\ddag} & 0.3034\textcolor{red}{\ddag} \\
                          & $\checkmark$ & & $\checkmark$ & 0.5556\textcolor{red}{\ddag} & 0.5291\textcolor{red}{\ddag} & 0.2579\textcolor{red}{\ddag} \\
                          & $\checkmark$ & $\checkmark$ & & 0.5778\textcolor{red}{\ddag} & 0.5562\textcolor{red}{\ddag} & 0.3027\textcolor{red}{\ddag} \\
                          & $\checkmark$ & & & 0.4222\textcolor{red}{\ddag} & 0.4321\textcolor{red}{\ddag} & 0.2229\textcolor{red}{\ddag} \\ \midrule
                            
\multirow{5}{*}{NATOPS}   & $\checkmark$ & $\checkmark$ & $\checkmark$ & 0.5778\textcolor{white}{\ddag} & 0.6114\textcolor{white}{\ddag} & 0.5898\textcolor{white}{\ddag} \\
                        & & $\checkmark$ & $\checkmark$ & 0.5778\textcolor{white}{-} & 0.6114\textcolor{white}{-} & 0.5898\textcolor{white}{-} \\
                          & $\checkmark$ & & $\checkmark$ & 0.5778\textcolor{white}{-} & 0.6114\textcolor{white}{-} & 0.5898\textcolor{white}{-} \\
                          & $\checkmark$ & $\checkmark$ & & 0.4763\textcolor{red}{\ddag} & 0.4678\textcolor{red}{\ddag} & 0.4433\textcolor{red}{\ddag} \\
                          & $\checkmark$ & & & 0.4463\textcolor{red}{\ddag} & 0.4612\textcolor{red}{\ddag} & 0.3925\textcolor{red}{\ddag} \\
\bottomrule
\end{tabular}
\end{table}

\section{Concluding Remarks}
This paper presents TFEC, a novel temporal-frequency enhanced contrastive learning framework for multivariate time-series clustering that mitigates augmentation-induced distortions and leverages latent cluster structures. The co-enhancement mechanism produces representations via adaptive frequency mixing while preserving temporal structure. A synergistic dual-path architecture improves cluster distribution and representation robustness through pseudo-label guided contrastive learning path and reconstruction adjustment path. Extensive evaluations show SOTA performance on six UEA datasets, with an average gain of 4.48\% on NMI metric over counterparts. As this work is precision-oriented, future directions include enhancing noise robustness and computational efficiency via lightweight enhancement.

\bibliographystyle{IEEEbib}
\bibliography{refs}

\end{document}